\documentclass[10pt,letterpaper]{article}

\usepackage{times}
\usepackage{epsfig}
\usepackage{graphicx}
\usepackage{amsmath}
\usepackage{amssymb}
\usepackage{bbm}
\usepackage[utf8]{inputenc}
\usepackage{soul}
\usepackage{graphicx}
\usepackage{booktabs}

\usepackage{amsthm}
\usepackage{mathtools}
\usepackage[margin=2cm]{geometry}

\usepackage{nips14submit_e}

\usepackage[pagebackref=true,breaklinks=true,letterpaper=true,colorlinks,bookmarks=false]{hyperref}

\usepackage[pagebackref=true,breaklinks=true,letterpaper=true,colorlinks,bookmarks=false]{hyperref}

\newcommand{\BEQ}{\begin{equation}}
\newcommand{\EEQ}{\end{equation}}
\DeclareMathOperator*{\argmax}{argmax}

\DeclareMathOperator{\Tr}{Tr}

\newcommand{\ph}{\phantom{abc}}

\DeclareMathAlphabet{\mathpzc}{OT1}{pzc}{m}{it}
\font\dsrom=dsrom10 scaled 1200
\newcommand{\indicator}[1]{\textrm{\dsrom{1}}_{#1}}

\nipsfinalcopy

\title{
    Semidefinite and Spectral Relaxations for Multi-Label Classification    \\
    \textcolor{red}{
        \small{
            \textnormal{
            }
        }
    }
}

\author
{
    R\'emi Lajugie \\
    INRIA, Sierra Project-team \\
    \'Ecole Normale Sup\'erieure, Paris\\
    \texttt{remi.lajugie@ens.fr}
    \And
    Piotr Bojanowski \\
    INRIA, Willow Project-team \\
    \'Ecole Normale Sup\'erieure, Paris\\
    \texttt{piotr.bojanowski@inria.fr}
    \AND
    Sylvain Arlot\\
    INRIA, Sierra Project-team \\
    \'Ecole Normale Sup\'erieure, Paris\\
    \texttt{sylvain.arlot@ens.fr}
    \And
    Francis Bach\\
    INRIA, Sierra Project-team \\
    \'Ecole Normale Sup\'erieure, Paris\\
    \texttt{francis.bach@inria.fr}
}

\begin{document}

\maketitle

\begin{abstract}
In this paper, we address the problem of multi-label classification. 
We consider linear classifiers and propose to learn a prior over the space of labels to directly leverage the performance of such methods. 
This prior takes the form of a quadratic function of the labels and permits to encode both attractive and repulsive relations between labels.
We cast this problem as a structured prediction one aiming at optimizing either the accuracies of the predictors or the $F_1$-score. 
This leads to an optimization problem closely related to the max-cut problem, which naturally leads to  semidefinite and spectral relaxations.
We show on standard datasets how such a general prior can improve the performances of multi-label techniques.
\end{abstract}

\section{Introduction}

Multi-label classification aims at predicting a set of labels for each data instance~\cite{Tsoumakas2007multi,Zhang2013review}. 
This setting is ubiquitous in real-world applications and for example can take the form of video or text tagging, where the goal is to assign instances to categories \cite{Joachims1998text}. 
For video, \cite{Xiao2010sun} proposes to consider the problem of labeling scenes, on which several objects appear.

One of the main difficulties of this problem lies in the fact that the space of potential labelings $\mathcal{Y}$ is exponentially bigger than the set of labels $\mathcal{V}$. 
Doing an exhaustive search over the space of labelings is thus not possible.
Moreover, contrary to the standard binary classification setting, the set $\mathcal{V}$ has a specific structure and one has to take it into account, especially when the number of labels is large. Indeed, imagine that we are given one classifier $f_v$ for each $v \in \mathcal{V}$, we would probably observe that some $f_v$ predict labels that are not actually present; for instance, in image tagging, if it is very likely to see a zebra and a lion on the same image, it is rather not probable to see a reindeer with a lion. A prior over labels could have, for instance, penalized the prediction of a reindeer together with the lion.
Incorporating structure into the label set can be done a priori by assuming labels are organized in a certain hierarchy \cite{Rousu2006kernel}; \cite{Hariharan2010large} incorporates a prior knowledge when training the classifiers, permitting to learn correlated classifiers. However this prior does not affect the way predictions are done.

Our goal is to learn such a prior over labels directly from data, at the same time that classifiers are learnt. This idea has already been tackled by \cite{Petterson2011submodular} who restricted their study to the specific case of incorporating \emph{positive affinities} between labels. We go beyond this approach and propose a model permitting to take into account affinities and \emph{incompatibilities} between labels.

\textbf{Related work.} 

A large part of the recent literature considers a moderately large set of labels $\mathcal{V}$ (order of hundreds) and a huge space of labelings $\mathcal{Y}$.
In this setting it is possible to learn specific classifiers for each label separately. 
One way to train such classifiers is the well-known one-versus-rest technique (a.k.a.~binary relevance technique \cite{Tsoumakas2007multi}).

Within this setting,  some approaches use the structured prediction framework \cite{Tsochantaridis2005large,Taskar2003maximum} as we do.
This corresponds to considering the task of prediction as being a task over the huge output space~$\mathcal{Y}$. 
\cite{Lampert2011maximum} has proposed to plug a model within a structured SVM, and considers the prior knowledge between labels as fixed a priori, whereas we aim at learning it.
They defined a proper loss and the corresponding loss-augmented decoding. 
The loss they used is called the ``max loss'' and is slightly related to the Hamming loss. 
This approach leads to an efficient loss-augmented decoding, and avoids an exhaustive search over the power set~$\mathcal{Y}$. 
Other approaches \cite{Dembczynski2013optimizing} considered the direct optimization of the $F_1$-score within a structured SVM.
Another part of the recent literature dealing with multi-label classification \cite{Bi2013efficient} considers the case where the space of labels $\mathcal{V}$ itself is huge.
In these papers, the goal is to use the fact that only few labels are present in an instance.
This allows to reduce the dimension of the prediction space and performing the labeling over a lower dimensional space.
The priors we propose here could be combined with these approaches.

\textbf{Contributions.}
Our contribution is four-fold: 
(1) we propose a model with priors for multi-label classification allowing attractive as well as repulsive weights, 
(2) we cast the learning of this model into the framework of structured prediction using either Hamming of $F_1$ losses and propose an approach for solving exactly the loss-augmented decoding using the $F_1$ loss,
(3) we propose semidefinite  and spectral relaxations to efficiently solve the resulting structured prediction problem,
(4) we show on real datasets how the learning of such a general prior can improve the multi-label prediction over the models where no prior is learnt or when only attractive weights are allowed.

\section{Structured Prediction for Multi-Label Classification}

In this section, we  review several ways to perform the multi-label classification task when a prior over the labels is fixed.
Decoding consists in assigning potentially several labels to a data point belonging to some feature space.
We then discuss how to learn the parameters of the predictive function.
For the rest of the paper we denote our feature space by \(\mathcal{X} \subset \mathbb{R}^d\).

\subsection{The multi-label classification problem problem}

Let us consider the set of possible labels $\mathcal{V}$ of cardinal $V$. 
We define the set of \emph{labelings}, as the set of binary vectors $\mathcal{Y} = \{-1,1\}^{V}$.
The set \(\mathcal{Y}\) is the one on which we perform our structured prediction.

Let us assume that for each possible label $v$, we are given a linear classifier parameterized by $ w_v \in \mathbb{R}^{d}$.
We denote by $W \in \mathbb{R}^{d \times V}$, the vertical concatenation of all the vectors \(w_v\).
In the multi-label setting, the decoding problem is:
\BEQ 
    \widehat{y}(x;W) \in \argmax_{y \in \{-1, 1\}^V} D(x, y ; W) := y^{\top} W^{\top} x.
\EEQ 
This is usually referred to as the binary relevance method for multi-label learning  \cite{Tsoumakas2007multi}. 

The aforementioned approach does not take into account any dependency between the different labels. 
A way to do so is to penalize the discriminative function by some penalty $F$ depending on the subset of predicted labels.
In our case, we propose to consider:
\BEQ \label{eq:ourDecoding}
    \widehat{y} \in \argmax_{y \in \{-1, 1\}^V} D(x,y ; W, F) := y^{\top} W^{\top} x - F(y).
\EEQ
However, not all functions \(F\) are admissible, so that   (\ref{eq:ourDecoding}) remains tractable since $|\mathcal{Y}|=2^{V}$.

A class of penalizations that are well-suited for our problem is the class of submodular functions \cite{Bach2013learning,Petterson2011submodular}. 
When $F$ is submodular, the decoding becomes the maximization of a supermodular function (maximization of a modular minus a submodular function).
This is known to be tractable (solvable in polynomial time in $V$).
\cite{Petterson2011submodular} has proposed to use a graph-cut based penalty.
This corresponds to $F(y) = y^{\top}A y - y^{\top} b$ where $b \in \mathbb{R}^{V}$ and  $A \in \mathbb{R}^{V \times V}$ is proportional to the Laplacian matrix of a graph. 
Intuitively, this corresponds to considering that labels are organized in a graph $G$ with non-negative weights, encoding \emph{attractive} affinities between the labels;
 the linear part of the prior $b$ corresponds to a prior over the frequencies of the classes.

For general weights, meaning that the matrix $A$ not only encodes affinities but also costs, the decoding task becomes as hard as solving a max-cut problem.
In Sec.~\ref{sec:relax} we review common convex relaxations permitting to obtain a good approximate solution in polynomial time.
Using a matrix $A$ with arbitrary entries, our decoding model becomes:
\begin{equation} 
    \label{eq:decoding}
    \widehat{y}(x; W, A, b) \in \argmax_{y \in \{-1, 1\}^V} D(x,y ; W, A, b) = \argmax_{y \in \{-1, 1\}^V} y^{\top} W^{\top} x + y^{\top} b- y^{\top} A y.
\end{equation}

\subsection{Learning the parameters \(W\), \(b\) and \(A\)}

In the previous section we have assumed that we are given \(V\) linear classifiers \(w_v \in \mathbb{R}^{d}\), a linear prior $b\in \mathbb{R}^{V}$ and a matrix \(A \in \mathbb{R}^{V\times V}\).
Thus, the discussed decoding problem  can be seen as being parameterized by \(W\), $b$ and \(A\).

Suppose that we are given \(N\) examples \((x_i, y_i) \in \mathcal{X} \times \mathcal{Y}, i=1,\dots, N\), and consider a loss function between two labelings \(\ell: \mathcal{Y}\times\mathcal{Y} \to \mathbb{R}_+\). 
Ideally, given this loss, we would like to minimize the following regularized empirical loss:

\begin{equation}\label{eq:empiricalLossMinimization}
\min_{W, A, b} \frac{1}{N} \sum_{i=1}^N \ell( \widehat{y}(x_i; W, A), y_i) + \lambda \Omega (W, A),
\end{equation}
where \(\Omega\) is a convex regularizer (typically a squared $\ell_2$-norm) over the parameter space.
This is a hard combinatorial problem that thus needs to be relaxed.
Following \cite{Tsochantaridis2005large,Taskar2003maximum}, we define the structural hinge loss \(H\) as:
\begin{equation}
	H(x_i, y_i, W, A, b) = \max_{y \in \{-1, 1\}^V} \left \{  \ell(y, y_i) + D(x_i, y; W, A, b) - D(x_i, y_i, W, A) \right \}.
\end{equation}
We estimate parameters \(W^*, b^{*}\) and \(A^*\) by solving the following problem:
\begin{equation}
\label{eq:structuredPredictionObjective}
    \min_{W, A, b} \ \frac{1}{N} \sum_{i=1}^N H(x_i, y_i; W, A, b) + \lambda \Omega(W, A, b).
\end{equation}

%%%%%%%%%%%%%%%%%%%%%%%%%%%%%%%%%%%%%%%%%%%%%%
%%% LOSSES FOR THE MULTI-LABEL PREDICITION %%%
%%%%%%%%%%%%%%%%%%%%%%%%%%%%%%%%%%%%%%%%%%%%%%

\section{Performance Measures and Losses for Multi-Label Tasks}
\label{sec:losses}
In order to set up the aforementioned problem, we need to define a proper loss function \(\ell\).

\textbf{Normalized Hamming loss.}
The simplest loss is based on accuracy, and is defined as:
\BEQ
a(y,y_i) = \frac{V+y^{\top}y_i}{2V} \in [0,1] \, .
\EEQ
The loss associated to accuracy is the so-called Hamming loss \cite{Hamming1950error,Zhang2013review}. 
It is defined as a linear function of the binary label vector $y$ by:
\begin{align}
    \ell(y,y')  &= \left \| \frac{1}{2\sqrt{V}}(\mathbf{1}-y) - \frac{1}{2\sqrt{V}}(\mathbf{1}-y_i) \right \|^{2}_2 \\
                &= \frac{1}{2V} \left ( V - {y_i}^{\top} y  \right ) = 1-a(y,y_i) \in [0,1] \,  ,
\end{align}
where $\mathbf{1}$ is the $V$-dimensional vector with ones.
This loss corresponds to the symmetric difference between two sets $A\Delta B = (A \cup B) \setminus (A \cap B)$.
Note also that, if we consider that not all the errors are equivalent, one can use a weighted Hamming loss instead.

\textbf{$F_1$ loss.}
A common choice in the multi-label learning literature is the $F_{\beta} -\text{score}$ loss \cite{Tsoumakas2007multi,Petterson2011submodular}. 
This loss is a function of precision and recall and has some important advantages over the Hamming loss.
In the common situations where each instance has only few labels among all the ones that are possible, the $F_{\beta}$ loss penalizes a lot the solution \((-1, \dots, -1)^{\top}\)  while the Hamming does not.

Precision and recall with respect to a training labeling $y_i \in \mathcal{Y}$ are defined respectively as: 
\[p(y, y_i)=\frac{(1+y_i)^{\top}(1+y)}{(1+y)^{\top}(1+y)}, \qquad r(y, y_i)=\frac{(1+y_i)^{\top}(1+y)}{(1+y_i)^{\top}(1+y_i)}.\] 
Then the general $F_{\beta}$ score is defined as, for every $\beta>0$,  
\begin{equation}
    F_{\beta}(y, y_i)=\frac{(1+\beta^{2}) \ p(y,y_i) \ r(y,y_i)}{\beta^{2} \ p(y, y_i) + r(y, y_i)} \in [0,1] \, .
\end{equation}
The most widely used is the $F_1$ score (which turns out to be the harmonic mean of precision and recall), and the associated loss is then $\ell(y, y_i) = 1-F_1(y, y_i)$. 
More precisely:
\begin{equation} 
    \ell(y, y_i) = \frac{V - y^{\top} y_i}{2V + y_i^{\top}\mathbf{1}+y^{\top}\mathbf{1}} \in [0,1] \, .
\end{equation}
Please note the non linear dependency of this loss in \(y\).

%%%%%%%%%%%%%%%%%%%%%%%%%%%%%%%
%%% LOSS AUGMENTED DECODING %%%
%%%%%%%%%%%%%%%%%%%%%%%%%%%%%%%

\section{Loss-Augmented Decoding}
\label{sec:DEC}
We propose to derive a structured-SVM-like optimization objective \cite{Tsochantaridis2005large}.
As mentioned earlier, we want to learn the parameters of our predictive function using annotated data.
Following the definition of $H$, we can write the complete optimization problem~(\ref{eq:structuredPredictionObjective}) as:
\begin{align}
    \label{eq:wholeCost}
    \min_{W, A,b} \ \frac{1}{N} \sum_{i=1}^N &  \left [ \max_{y \in \{-1, 1\}^V} \left \{ \ell(y_i, y) + y^{\top} W^{\top} x_i + y^{\top} b - y^{\top} A y \right \} - y_i^{\top} W^{\top} x_i - y_i^{\top} b + y_i^{\top} A y_i \right ] \nonumber \\ 
                  & \qquad \hspace*{6cm} +\frac{\lambda_W}{2} \|W\|_2^2 + \frac{\lambda_A}{2} \|A\|_2^2.
\end{align}
%However, we need to detail the loss \(\ell\) since in some cases this can yield a more complex optimization problem.

\textbf{Using the Hamming loss.}
If we use the Hamming loss for \(\ell\), then \(\ell(y_i, y) = \frac{1}{2V} \left ( V - y^{\top} y_i \right )\).
Our optimization problem can be re-written as follows:
\begin{align}
    \label{eq:HammingCost}
    \min_{W, A,b} \ \frac{1}{N} \sum_{i=1}^N & \left [ \max_{y \in \{-1, 1\}^V} \left \{ y^{\top} \left ( W^{\top} x_i + b - \frac{1}{2V} y_i \right ) - y^{\top} A y \right \} - y_i^{\top} W^{\top} x_i - y_i^{\top} b + y_i^{\top} A y_i \right ] \nonumber \\
                  & \qquad \hspace*{6cm}  + \frac{\lambda_W}{2} \|W\|_2^2 + \frac{\lambda_A}{2} \|A\|_2^2.
\end{align}
Note that the objective function of the optimization  is jointly convex but not smooth.

\textbf{Using the $F_1$ loss.}
If in turn we decide to use the $F_1$ loss, the proposed problem is harder because of the vector \(y\) in the denominator.
To cope with this issue, we can split the set \(\mathcal{Y}\) into \( (V+1)\) subsets.
We define the set \(\mathcal{Y}_k\) as the set of labelings such that \(k\) entries are positive:
\[
    \forall k \in \{0, \ldots V\}, \ \ \mathcal{Y}_k = \left \{ y \in \{-1, 1\}^V \ , \ y^{\top} \mathbf{1} = 2 k - V \right \}.
\]
As is often done when optimizing the $F_1$ score, which is a contingency-table based loss \cite{Joachims2005support}, we can divide the initial problem into $V+1$ subproblems by replacing \(y^{\top} \mathbf{1}\) by \(2k-V\) as follows: 
\begin{equation} 
    \label{eq:subprob}
    \max_{k \in \{0, \dots, V\}}  \left [ \frac{V}{V + y_i \mathbf{1} + 2k} + \max_{y \in \mathcal{Y}_k} \left \{ y^{\top} \left ( \frac{y_i}{V+y_i \mathbf{1} + 2k } + W^{\top} x_i + b \right ) - y^{\top} A y \right \} \right ] .
\end{equation} 
The   problems of Eq.~\eqref{eq:HammingCost}--\eqref{eq:subprob} above assume that we are able to solve quadratic optimization problems for $y \in \mathcal{Y}$. 
\cite{Petterson2011submodular} proposes a greedy approximate algorithm for solving this type of problems in the specific case where off diagonal entries of the prior $A$ are negative.

In the following section, we propose relaxations of these problems leading to a tractable loss-augmented decoding with no restriction over the matrix $A$.

%%%%%%%%%%%%%%%%%%%%
%%% OPTIMIZATION %%%
%%%%%%%%%%%%%%%%%%%%

\section{Optimization in $y$ }
\label{sec:optim-y-z}

So far, we have written three problems that we are not able to solve efficiently.
The first one was the general decoding of Eq.~\eqref{eq:decoding}.
The other ones were the subproblems of Eq.~\eqref{eq:HammingCost} and Eq.~\eqref{eq:subprob}.
All of these are quadratic boolean optimization problems and are closely linked to the max-cut problem (see, e.g., \cite[Sec.~5.1.5]{Boyd2004convex}).
These can be written in the canonical form as follows:
\begin{equation} 
    \label{eq:canonical}
    \max_{\substack{u \in \{-1,1\}^{V}\\ L(u) = 0}} u^{\top}b - u^{\top} A u,
\end{equation} 
where $A \in \mathbb{R}^{V \times V} $, $b \in \mathbb{R}^V$ and $L$ is an affine function. 

Note the presence of the additional constraint \(L(u) = 0\).
This additional equation is only needed for the problem mentioned in Eq.~(\ref{eq:subprob}).
For the two other problems, one can simply ignore it.
Eq.~(\ref{eq:canonical}) allows us to tackle three problems in a unified framework.
In the next section we discuss two relaxations to this problem.
First we describe the standard SDP relaxation.
We then present how to cast this optimization problem as a spectral problem.

\subsection{Classical semidefinite relaxation for max-cut}
\label{sec:relax}

The family of problems presented in Eq.~(\ref{eq:canonical}) is known as the two-way partitioning problems. 
They are a generalization of max-cut, with potentially negative entries in $A$.
Also, they contain an extra linear term (see Sec.~5.1.5 of \cite{Boyd2004convex}) and potential constraints over the domain.

There exists a classical semidefinite relaxation. Following \cite{Boyd2004convex,Aspremont2003relaxations}, we use a similar relaxation to the one used by \cite{Goemans1995improved} to approximate the max-cut problem.
We introduce a new variable $U = u u^{\top} \in \mathbb{R}^{V \times V}$.
Using this notation we can re-write the term \(u^{\top} A u\) as \(\text{Tr} \left ( A U \right )\).
Then using a set of constraints that is equivalent to \(U = u u^{\top}\) the problem (\ref{eq:canonical}) can be re-written as:
\begin{equation} 
	\max_{\substack{u \in \{-1, 1\}^V \\ U \in \mathbb{R}^{V \times V}}} \quad  u^{\top}b - \text{Tr}(AU) \qquad
\text{such that} \qquad
\begin{cases}
    	\text{Diag} (U) = \mathbf{1}, \\
	 \text{Rank}(U) = 1, \\
         	U \succeq uu^{\top}, \\
          L(u) = 0 .
          \end{cases}
\end{equation}
Following \cite{Boyd2004convex}, the convex relaxation of this problem is obtained by removing the rank constraint.
We define \(L\) as the affine function $L(u) = u^{\top} \alpha - \beta$ where $\alpha \in \mathbb{R}^V$ and $\beta \in \mathbb{R}$.
We use the Schur complement trick (see, e.g., \cite{Boyd2004convex}) and define the matrix $M$ as:
\[
    M = 
\begin{pmatrix} 
    U   & u \\ 
    u^{\top} & 1
\end{pmatrix}.
\]
Using $e_V$, the vector with all coordinates equal to zero except the last one, our relaxation of (\ref{eq:canonical}) can be re-written as:
\begin{align} 
	\label{eq:sdpRelax}
	\max_{M \in \mathbb{R}^{V \times V}} \quad   \text{Tr} 
	\left [ 
                            M 
                            \begin{pmatrix}
                                -A              & \frac{1}{2}b\\ 
                                \frac{1}{2}b    & 0
                            \end{pmatrix}
	\right ] \qquad
	\text{such that} \qquad  
	\begin{cases}
		\text{Diag}(M) = \mathbf{1},\\
	        M \succeq 0, \\
		\alpha^{\top} M e_V = \beta.
	\end{cases}
\end{align}
Problem (\ref{eq:sdpRelax}) can be solved using any standard convex optimization solver at least for small $V$ ($<100$).
When $V$ is large, one can use specific techniques relying explicitly on the fact the solution is expected to be low-rank (see, e.g., \cite{Journee2010low} and references therein).

\textbf{Rounding scheme}

At test time, we follow \cite{Boyd2004convex} to round the relaxed solution, \emph{i.e.}, get back to some admissible solution of (\ref{eq:canonical}).
We notice that at the optimum $(u,U)$ of Eq.~(\ref{eq:sdpRelax}), $U \succeq u^{\top}u$ implies that $U-uu^{\top}$ is a covariance matrix. 
Therefore, we simply sample several $v\sim \mathcal{N}\big(u, U -uu^{\top}\big)$ from a normal distribution, round the solution by taking the signs and choose the best one in terms of the objective function.
This procedure leads to good feasible points in our experiments.

\subsection{Spectral relaxation}
\label{sec:spectral}

The generic problem in Eq.~(\ref{eq:canonical}) can be rewritten by replacing the integrality constraint \(u \in \{-1, 1\}^V\) with a quadratic equality $u^{\top}u = V$.
Please note this makes the problem non-convex.
Using the same expression for \(L(u)\) as in the previous section leads to the following optimization problem:
\begin{equation}
\label{eq:spectralRelax}
\max_{u \in \mathbb{R}^V} \quad u^{\top} b - u^{\top} A u \qquad
\text{such that} 	\qquad
\begin{cases}			
	u^{\top}u= V \\
	u^{\top} \alpha = \beta.
\end{cases}
\end{equation}
We deal with the linear constraint by dualizing it, yielding the following problem:
\begin{equation}
    \label{eq:spectralDualized}
    \min_{\mu \in \mathbb{R}} \bigg[ \mu \beta + \max_{\substack{u \in \mathbb{R}^V \\ u^{\top} u = V}} u^{\top} \left ( b - \mu \alpha \right ) - u^{\top} A u \bigg] .
\end{equation}
This can be solved by performing a binary search over \(\mu\).

The inner loop problem is classical in optimization, in particular in trust-region methods \cite{Forsythe1965stationary,Spjotvoll1972note}.
It reduces---using the Lagrange multiplier technique---to solving a quadratic eigenvalue problem \cite{Tisseur2001quadratic}.
Solving the inner loop problem of Eq.~\eqref{eq:spectralDualized} (with nonzero $b$) is equivalent to finding the minimal eigenvalue of the quadratic eigenvalue problem:
\BEQ 
\label{eq:quadraticEigenvalueProblem}
\textstyle \big(\lambda^2 I - 2 \lambda A + A^2 - \frac{1}{4V} (b - \mu \alpha) (b - \mu \alpha )^{\top} \big) u= 0,
\EEQ
where $I$ denotes the $V \times V$ identity matrix. The   problem above is solved efficiently by performing the SVD of the matrix $S$:
\begin{equation}
S = 
\begin{pmatrix}
A& -I \\
-\frac{1}{4V} (b- \mu \alpha)  (b - \mu \alpha)^{\top} & A
\end{pmatrix}.
\end{equation}
Once this has been solved, we get the desired solution by taking $u =\frac{1}{2} (A - \lambda I)^{-1} (b - \mu \alpha)$, where $\lambda$ is the smallest non-zero eigenvalue of $S$.

Note that when optimizing the Hamming loss, we get rid of the constraint \(L(u)=0\).
In that case we can set \(\mu = 0\) and solve the inner loop problem only once.

\subsection{Cheaper (but still efficient) solution for the spectral relaxation}

In this section we present an other way to deal with the spectral relaxation, inspired by \cite{Gander1989constrained}. The proposed method is more efficient computationnally than the one of the previous section since it does not involve solving the binary search problem over the Lagrange multiplier $\mu$.

We start from the problem of Eq.~\eqref{eq:spectralRelax}.
By the change of variables $v = \begin{pmatrix}
u \\ 1
\end{pmatrix}$ and $B = \begin{pmatrix}
-A & b/2 \\ b/2 & 0
\end{pmatrix}$ and by introducing $D = \begin{pmatrix}
I & 0 \\ 0 & 0
\end{pmatrix}$ ($I$ is the $V$ dimensional identity matrix) we can write the problem as:

\begin{equation}
\label{eq:spectralRelaxCompact}
\max_{v \in \mathbb{R}^{V+1}} \quad v^{\top} B v \qquad
\text{such that} 	\qquad
\begin{cases}			
	v^{\top}Dv= V \\
	v^{\top} \begin{pmatrix}
	\alpha & 0 \\ 0 & 1
	\end{pmatrix} =
	\begin{pmatrix}
	\beta \\
	1
\end{pmatrix}.
\end{cases}
\end{equation}

Following \cite{Gander1989constrained}, let us simply introduce the QR factorization of the matrix $\begin{pmatrix}
	\alpha & 0 \\ 0 & 1
	\end{pmatrix} = QR$, where $Q \in \mathbb{R}^{V+1 \times V+1}$ is an orthogonal matrix and $R \in \mathbb{R}^{V+1 \times 2}$.
Let us now introduce $U = \begin{pmatrix}
U_1 \\ U_2
\end{pmatrix}
=Q^TV$. $U_1 \in \mathbb{R}^2$ and $U_2 \in \mathbb{R}^{V-1}$

Eq.~\eqref{eq:spectralRelaxCompact} can be rewritten as:

\begin{equation}
\label{eq:spectralRelaxCompact2}
\max_{U \in \mathbb{R}^{V+1}} \quad U^{\top} Q^TBQ U \qquad
\text{such that} 	\qquad
\begin{cases}			
	U^{\top}DU= V \\
	U_1^{\top} R =
	\begin{pmatrix}
	\beta \\
	1
\end{pmatrix}.
\end{cases}
\end{equation}

Note that the last constraint permit to fix the variable $U_1^{\top} = R^{-1}\begin{pmatrix}
	\beta \\
	1
\end{pmatrix}.$ With a slight abuse of notation, $R^{-1}$ corresponds to the inverse of the rotation part.
Let us define $Q^TBQ = \begin{pmatrix}
\Delta & \Gamma/2 \\
\Gamma^{\top}/2 & C
\end{pmatrix},
$ with $\Delta \in \mathbb{R}^{2\times 2}$ , $\Gamma \in \mathbb{R}^{T \times 2}$ and $C \in \mathbb{R}^{(V-1) \times (V-1)}$.
Using the previous notations, we get:
$U^{\top} Q^{\top}BQ U = U_2^T C U_2 + U_1^{\top} \Gamma U_2 +U_1^{\top}\Delta U_1$.

Let us also introduce $S = V - U_1^TU_1$
Since $U_1$ is not entirely determined this problem is equivalent to:

\begin{equation}
\max_{U_2 \in \mathbb{R}^{V-1}} \quad U_2^{\top} C U_2 + R^T\begin{pmatrix}
	\beta \\
	1
\end{pmatrix}\Gamma U_2 \qquad
\text{such that} 	\qquad
\begin{cases}			
	U_2^{\top}DU_2 = S \\
\end{cases}.
\end{equation}
Note that we slightly abuse of notations with $D$ being restricted to its last components.

\subsection{Links with graph-cuts}

The min-cut problem can be written as an optimization problem through the following equation:
\begin{equation}
    \min_{z \in \{0, 1\}^V} \sum_{j=1}^V \sum_{i=1}^{j-1} C_{i,j} |z_i - z_j| + c^{\top} z,
\end{equation}
where $C \in \mathbb{R}_+^{V \times V}$ and \(c \in \mathbb{R}^V\).
By making the change of variables \(z = \frac{y+1}{2}\) and carrying on some calculations, we get the following equivalent program:
\begin{equation}
	\min_{y \in \{-1, 1\}^V} 2 y^{\top} c - y^{\top} C y.
\end{equation}

Therefore, when the matrix \(A\) has negative off-diagonal entries, the problem formulated in Eq.~(\ref{eq:canonical}) can be solved using min-cut / max-flow.
We can use standard min-cut / max-flow toolboxes by providing the matrix \(C=-A\) and \(c = \frac{1}{2} b\).

When optimizing the $F_1$ loss,   note that we can use the same dualization for the constraint \(L(u)=0\).
We proceed exactly as with the spectral relaxation except that the inner loop is solved with min-cut / max-flow. 

\subsection{Solving the F1 loss augmented decoding for negative $A$}

In this section, we show how we can solve the constrained problem by relating it to the well-studied total variation denoising problem~\cite{Chambolle2009total,Bach2013learning}.
Note that, contrary to \cite{Petterson2011submodular}, in this section, we deal with the cardinality constraint \emph{exactly} and we do not use any approximation algorithm in this specific case. We just use total variation minimization algorithm to perform the constrained minimization.

Here we consider that the constraint of Eq.~\eqref{eq:canonical} is simply a cardinality constraint, namely that it is of the form $u^{\top}c = \alpha$ for a certain $\alpha \in \{1\ldots V\}$ and $c$ is the $V$ dimensional vector composed of ones.

Now, we dualize this equality constraint by introducing the associated Lagrange multiplier. This yields the following problem:

\begin{align}
\label{eq:graph-cutDualized}
\max_{\mu \in \mathbb{R}}\min_{u \in \{-1, 1\}} u^{\top}Au - u^{\top}b + \mu (\alpha - u^{\top}c)
\end{align}

Equivalentally, by considering the variable $z \in \{0,1\}^V$ we get the following problem:

\begin{align}
\label{eq:graph-cutparametric}
\max_{\mu \in \mathbb{R}} \mu(\alpha - V) \min_{z \in \{0,1\}^V} 4 z^{\top} A z - z^{\top}(4Ac + 2 b) + \mu z^{\top}c.
\end{align}

The problem of Eq.\eqref{eq:graph-cutparametric} is a separable submodular optimization problem \cite{Bach2013learning}. Thus solving it can be done by considering the associated proximal problem.
More precisely, if we introduce the Choquet integral of the cut $J(u)$ (often referred to the ``co-area formula'' \cite{Chambolle2009total} for the specific case of cut functions or Lovasz extension for submodular functions), the generic proximal problem associated to any cut problem is:

\begin{align}
\label{eq:proximalGraph-cut}
\min_{u \in \mathbb{R}^V} \frac{1}{2}\|u - g\|^2_2 + J(u).
\end{align}

where $g$ in our case is exactly $4Ac + 2 b$.

This problem is the well known total variation denoising problem. There exists several efficient algorithms to deal with it, especially the ones relying on parametric max-flow techniques. 
Once problem\eqref{eq:proximalGraph-cut} has been solved and that we recovered its (unique if $\alpha$ is positive) solution $u^*$, we get all the candidates for being a solution of \eqref{eq:graph-cutDualized} by considering the different $\indicator{u\geq-\mu}$. Then, we just have to compute the associated objective values and select the optimal one.

\section{Optimization in $W$ and $A$}

We optimize our cost function in Eq.~(\ref{eq:wholeCost}) with stochastic subgradient descent.
When we relax the inner optimization problem in \(y\in \mathcal{Y}\) we implicitly modify the cost function.
Therefore we have to be careful when computing the subgradients.

In this section we provide the derivations in one specific case. 
The details for the other cases can be found in the supplementary material.
When using the Hamming loss and the SDP relaxation, our cost function becomes,
with $\mathcal{U} = \{(U, u), U \in \mathbb{R}^{V \times V}, u \in \mathbb{R}^V, U \succeq u^{\top}u, \text{Diag}(U) = \mathbf{1}_V\}$:
\begin{align}
    \min_{W, A, b} \ \frac{1}{N} \sum_{i=1}^N 
    & \left [ \max_{u, U \in \mathcal{U}} \left \{ u^{\top} \left ( W^{\top} x_i + b - \frac{1}{2V} y_i \right ) - \Tr(AU) \right \} - y_i^{\top} W^{\top} x_i - y_i^{\top} b + y_i^{\top} A y_i \right ] \nonumber \\[-.25cm]
    & \qquad     \hspace*{6cm}  + \frac{\lambda_W}{2} \|W\|_2^2 +  \frac{\lambda_A}{2} \|A\|_2^2.
\end{align}
To obtain the subgradients, we first solve the relaxed loss-augmented inference.
Using the obtained \(u\) and \(U\), we compute the subgradients in $W$ and $A$ as follows: 
\begin{align}
\label{eq:subgradients}
\partial_W g (W, A) &= \textstyle \lambda_W W + \frac{1}{N} \sum^{N}_{i=1} x_i (u-y_i)^{\top}, \\[-.1cm]
\partial_b g (W, A) &= \textstyle\frac{1}{N} \sum^{N}_{i=1} (u - y_i), \\[-.1cm]
\partial_A g (W, A) &= \textstyle \lambda_A A + \frac{1}{N} \sum^{N}_{i=1} - U + y_iy_i^{\top}.
\end{align}

\section{Experimental Evaluation}

We now validate the proposed approach on standard benchmarks.
We compare our implementation to \cite{Petterson2011submodular} and to the one-versus-rest model (OvR).
The code corresponding to the described method will be made publicly available.
In this experimental section we first describe the used datasets and discuss the baselines to which we compare.

\textbf{Datasets.}
We validate our approach on four datasets.
Following \cite{Petterson2011submodular}, we picked our datasets from the \emph{mulan}\footnote{http://mulan.sourceforge.net/datasets.html} repository.
We picked the \emph{yeast} \cite{Elisseeff01akernel}, \emph{enron}, \emph{medical} \cite{Pestian2007shared} and \emph{bibtex} \cite{Katakis08multilabeltext} datasets.
The datasets are of various sizes and natures: \emph{yeast} only has 14 labels while \emph{bibtex} has 159.
All of them also present different challenges (different structures, label concurrence patterns, etc.).

These datasets are given with a \emph{train} / \emph{test} split.
We further split the training set to generate a validation set.
We select all relevant parameters by plain validation on this set.
We report all performances on the actual test set as given in the dataset.
Caracteristics of these datasets are given in Table~\ref{table:charac}.

\begin{table}[ht]
    \centering
    \begin{tabular}{rrrr}
        \toprule
                & Instances               & Features       & Labels \\
        \midrule
        yeast   & 2417            & 103                 & 14 \\
        enron   & 1702            & 1001                & 53 \\
        medical & 978             & 1449                & 45 \\
        bibtex  & 7395            & 1836                & 159 \\
        \bottomrule
    \end{tabular}
    
    \vspace*{-.125cm}
   
    \caption{ Standard characteristics of the datasets used. \label{table:charac}
    }

\end{table}

\textbf{One-versus-rest results.}
In Table~\ref{tab:svmPerf} we report the performance of a one-versus-rest model for all the datasets.
For every label, we train a linear classifier using a standard SVM toolbox \cite{Fan2008liblinear}.
We select the hyper-parameters by validation on a held-out part of the training set.
We compare three criteria for choosing the optimal set of regularization parameters.
We can either select a common regularization parameter for all classes (``Single \(\lambda\)'' column), chosen with the Hamming loss (which decouples over classes), or one per class (``Multiple \(\lambda\)'' column).
When choosing a common $\lambda$ for all classes, one can choose it according to the $F_1$ or Hamming loss on the validation set.

% FB: what does this mean?
% When choosing one \(\lambda\) per class, both losses are equivalent.

\begin{table}[ht]
    \centering
    \begin{tabular}{rrrrr}
        \toprule
                & \multicolumn{2}{c}{Single $\lambda$}  &\ph& Multiple $\lambda$ \\
        \cmidrule{2-3} \cmidrule{5-5}
                & $F_1$               & Hamming            &   & Hamming \\
        \midrule
        yeast   & 0.39            & 0.40              &   & 0.54 \\
        enron   & 0.48            & 0.49              &   & 0.46 \\
        medical & 0.29            & 0.29              &   & 0.28 \\
        bibtex  & 0.61            & 0.66              &   & 0.66 \\
        \bottomrule
    \end{tabular}
    
      \vspace*{-.125cm}

    \caption{Linear SVM performance on the considered datasets. We report the average $F_1$ loss for various schemes for choosing the regularization parameter $\lambda$.}
    \label{tab:svmPerf}
\end{table}

Table~\ref{tab:svmPerf} shows that it is sometimes important to use the relevant loss as a criterion to select hyperparameter. In our experiments, this becomes more and more important as the size of the label set increases and thus as discussed in Sec.~\ref{sec:losses} the Hamming loss behaves more and more differently from the $F1$ loss.

One would also expect that picking one parameter per label would lead to better performance.
But the benefits from selecting a specific parameter per class is offset by the fact that one cannot use the $F_1$ loss in this case. In all our remaining simulations, we use a single $\lambda$ for all classes.

\textbf{Our model and comparison to \cite{Petterson2011submodular}.}
We run our algorithm---with $\ell$ equal to the Hamming loss---on all four datasets and compare to the available implementation of \cite{Petterson2011submodular}. 
For all methods we select all hyper-parameters based on the performance in terms of $F_1$ loss on the validation set.
Because of the challenging number of labels for \emph{bibtex}, we were  able to run neither the code from \cite{Petterson2011submodular}, nor the SDP, in reasonable time.

\begin{table}[ht]
    \centering
    \begin{tabular}{rrrrrrrrrrrr}
        \toprule
        &&&&& \multicolumn{3}{c}{SDP} && \multicolumn{3}{c}{Spectral} \\
        \cmidrule{6-8} \cmidrule{10-12}
        && OvR   & \cite{Petterson2011submodular}  & MC   & $A\leqslant0$ & $A\geqslant0$ & Any $A$ && $A\leqslant0$ & $A\geqslant0$ & Any $A$ \\
        \midrule                                                       
        yeast   && 0.39 &  \textbf{0.36}            & 0.40     & 0.40    & 0.39  & 0.39  && 0.39 & 0.37  & 0.37 \\
        enron   && 0.48 &  \textbf{0.45}            & 0.47     & 0.47    & 0.47  & \textbf{0.45}  && 0.48 & 0.49  & 0.49 \\
        medical && 0.29 &  0.33            & 0.29     & 0.31    & 0.29  & 0.24  && 0.30 & \textbf{0.21}  & 0.24 \\ 
        bibtex  && 0.61 &  N/A             & 0.61     & N/A     & N/A   & N/A   && 0.62 & \textbf{0.57}  & 0.60 \\
        \bottomrule
    \end{tabular}
  
      \vspace*{-.125cm}
  
    \caption{
        Comparison between \cite{Petterson2011submodular} and different variants of our method. 
        OvR denotes the one-versus-rest approach.
        MC is our algorithm with the inner loop being solved using min-cut / max-flow.
        SDP is the semidefinite relaxation of the inner loop.
        Spectral is the spectral relaxation of the inner loop.
    }
    \label{tab:main-tab}
\end{table}

Table~\ref{tab:main-tab} compares the one-versus-rest approach, the approach described in \cite{Petterson2011submodular} and variants of our method. 
We compare the two relaxations we proposed while optimizing the Hamming loss.
Please recall that the min-cut (MC) solution implies that $A \leqslant 0$ (non-positive entries).

When $A \leqslant 0$, we can measure the tightness of the proposed relaxations. We see that the various relaxations, SDP then spectral, do not degrade performances over the exact approach MC (which cannot be run for general $A$).

We also notice that using a negative matrix \(A\) is a strong limitation.
The performance observed when \(A\) is unconstrained or non-negative is better.
This motivates our formulation and shows that repulsive weights between labels are relevant.

\begin{table}[ht] \label{table:HammingvsF1}
    \centering
    \begin{tabular}{rrrrrrrrr}
        \toprule
        && OvR      & \cite{Petterson2011submodular} & Our Hamming   & Our $F_1$ \\
        \midrule                                                       
        yeast   && 0.39     & 0.37              & 0.43          & 0.43 \\
        enron   && 0.48     & 0.45              & 0.47          & 0.47 \\
        medical && 0.29     & 0.33              & 0.28          & 0.28 \\
        bibtex  && 0.61     & 0.58              & 0.60          & 0.60 \\
        \bottomrule
    \end{tabular}
    
    \vspace*{-.125cm}
    
    \caption{Comparison of $F_1$ losses when optimizing the $F_1$ loss versus the Hamming loss.}
\end{table}

\textbf{The Hamming loss and the $F_1$ loss.}
In this experiment we do not make use of the quadratic prior, so \(A=0\).
Table~\ref{table:HammingvsF1} gives the $F_1$ loss we obtain by optimizing either the $F_1$ loss or the Hamming loss. 
We compare the implementation of the $F_1$ score minimization in \cite{Petterson2011submodular} (carried out using a greedy technique).
In that table, ``Our $F_1$'' is our own implementation of the support vector technique for $F_1$-loss \cite{Joachims2005support} using the optimization described in Section~\ref{sec:DEC}.
This is an exact optimization technique. 
We also report the results obtained by training SVMs, using the one-versus-rest scheme.
It appears that, on these standard datasets ($V \approx 10-50$), optimizing the $F_1$ loss does not yield better performances than optimizing the Hamming loss.

\section{Conclusion}

We have proposed a framework to learn a prior for improving the performances of multi-label classification tasks. This prior takes the form of a quadratic function over the space of labels and incorporates both affinities and negative affinities. Existing work \cite{Petterson2011submodular} only takes into account positive affinities between labels.
We provide semidefinite and spectral relaxations of the learning problem, yielding to an efficient optimization scheme. In particular the spectral relaxation permits to deal computationally with  datasets rather large ($V>150$) whereas existing algorithms cannot (since the loss-augmented decoding problems have to solved many times).

It would be interesting to see how it is possible to leverage the range of applicability of the semidefinite relaxations which is, for now, limited to multi-label problems for which $V$ is of the order of hundreds. 
To that extent, we could use techniques from matrix optimization theory, taking into account for the fact that the solution we aim at finding has low rank \cite{Journee2010low}.

\bibliographystyle{plain}
\bibliography{biblio}
\renewcommand{\thesection}{\Alph{section}}

\end{document}